\documentclass[11pt,a4paper]{article}
\usepackage[hyperref]{acl2021}
\usepackage{times}
\usepackage{latexsym}
\usepackage{eurosym}
\usepackage{tikz-dependency}
\usepackage{booktabs}
\usepackage{multirow}
\usepackage{xspace}
\usepackage{xcolor}
\usepackage{hyperref}
\usepackage{graphicx}
\usepackage{subcaption}
\usepackage{relsize}
\usepackage{amsmath}
\usepackage{tabularx}
\usepackage[htt]{hyphenat}
\usepackage{arydshln}

\usepackage{microtype}

\usepackage{listings}
\usepackage{xcolor}

\definecolor{codegreen}{rgb}{0,0.6,0}
\definecolor{codegray}{rgb}{0.5,0.5,0.5}
\definecolor{codepurple}{rgb}{0.58,0,0.82}
\definecolor{backcolour}{rgb}{0.95,0.95,0.92}

\lstdefinestyle{mystyle}{
    backgroundcolor=\color{backcolour},   
    commentstyle=\color{codegreen},
    keywordstyle=\color{magenta},
    stringstyle=\color{codepurple},
    basicstyle=\ttfamily\footnotesize,
    breakatwhitespace=false,         
    breaklines=true,                 
    captionpos=b,                    
    keepspaces=true,                 
    showspaces=false,                
    showstringspaces=false,
    showtabs=false,                  
    tabsize=1
}
\aclfinalcopy 

\newcommand\skweak{{\small \textsf{skweak}}}
\renewcommand{\UrlFont}{\sffamily\small}

\title{\textsf{skweak}: Weak Supervision Made Easy for NLP}

\author{Pierre Lison \\
Norwegian Computing Center \\ Oslo, Norway \\
  {\normalsize \texttt{plison@nr.no}} \\ \And
  Jeremy Barnes \\
  \ \ \ Language Technology Group \\ University of Oslo \\
  {\normalsize \texttt{jeremycb@ifi.uio.no}} \\ \And
   \ \ Aliaksandr Hubin \\
  Department of Mathematics \\ University of Oslo \\
  {\normalsize \texttt{aliaksah@math.uio.no}} \\}

\date{}

\begin{document}
\maketitle
\begin{abstract}
    We present \skweak{}, a versatile, Python-based software toolkit enabling NLP developers to apply \textit{weak supervision} to a wide range of NLP tasks. Weak supervision is an emerging machine learning paradigm based on a simple idea: instead of labelling data points by hand, we use \textit{labelling functions} derived from domain knowledge to automatically obtain annotations for a given dataset. The resulting labels are then aggregated with a generative model that estimates the accuracy (and possible confusions) of each labelling function.  \\
    The \skweak{} toolkit makes it easy to implement a large spectrum of labelling functions (such as heuristics, gazetteers, neural models or linguistic constraints) on text data, apply them on a corpus, and aggregate their results in a fully unsupervised fashion. \skweak{} is especially designed to facilitate the use of weak supervision for NLP tasks such as text classification and sequence labelling. We illustrate the use of \skweak{} for NER and sentiment analysis. \skweak{} is released under an open-source license and is available at: \\ \url{https://github.com/NorskRegnesentral/skweak}
    
\end{abstract}

\section{Introduction}

Despite ever-increasing volumes of text documents available online, labelled data remains a scarce resource in many practical NLP scenarios.  This scarcity is especially acute when dealing with resource-poor languages and/or uncommon textual domains. This scarcity is also common in industry-driven NLP projects using domain-specific labels without pre-existing datasets. Large pre-trained language models and transfer learning  \cite{peters-etal-2018-deep,peters-etal-2019-tune,lauscher-etal-2020-zero} can to some extent alleviate this need for labelled data, by making it possible to reuse generic language representations instead of learning models from scratch. However, except for zero-shot learning approaches \cite{artetxe2019tacl,barnes2019-projecting,pires-etal-2019-multilingual}, they still require some amounts of labelled data from the target domain to fine-tune the neural models to the task at hand. 

\begin{figure}[t!]
    \centering
    \includegraphics[trim=5 3 3 3,clip, width=0.43\textwidth]{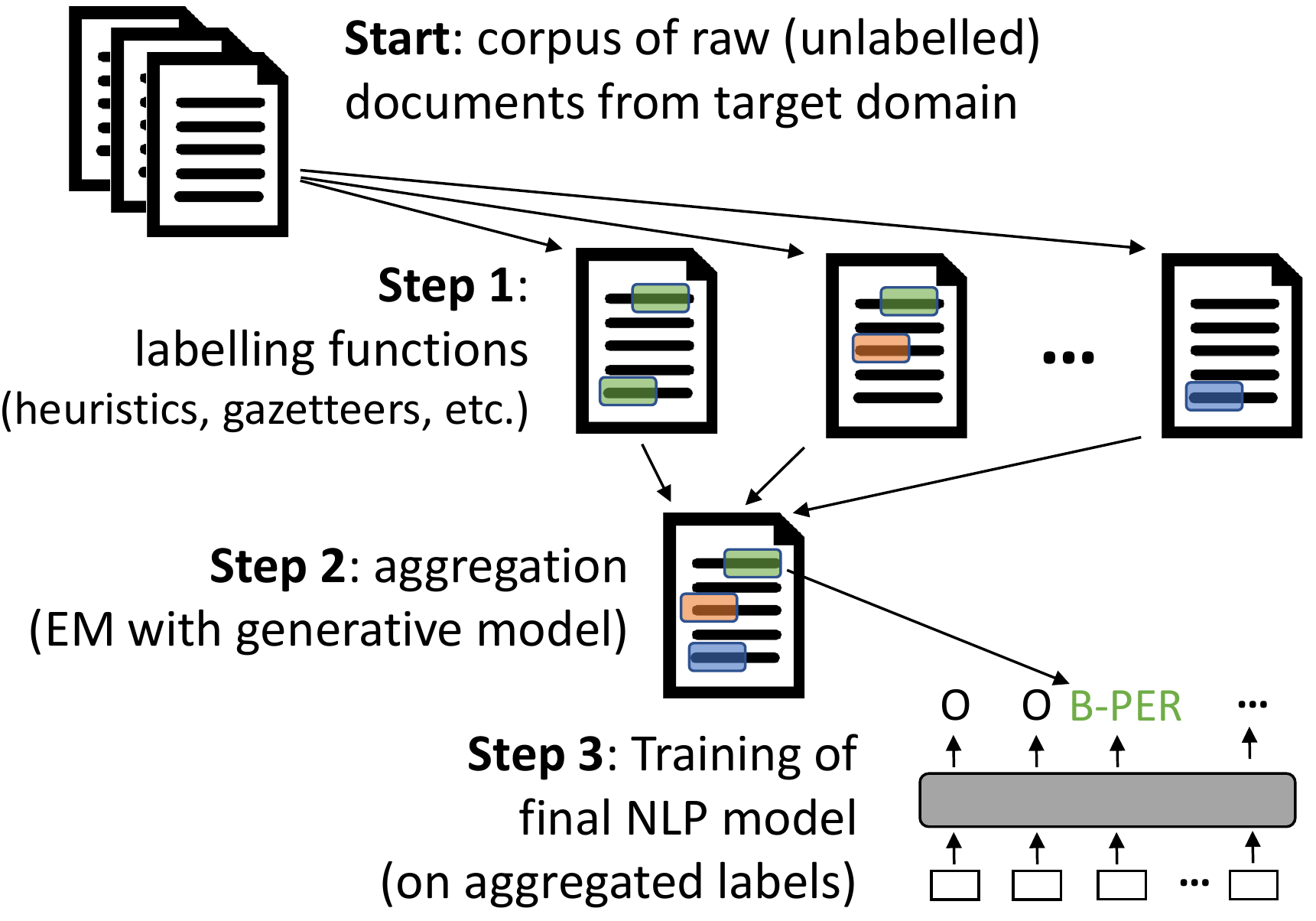}
    \vspace{2mm}
    \caption{General overview of \skweak{}: labelling functions are first applied on a collection of texts (step 1) and their results are then aggregated  (step 2). A discriminative model is finally trained on those aggregated labels (step 3). The process is illustrated here for NER, but \skweak{} can be applied to any type of sequence labelling or classification task.}
    \vspace{-2mm}
    \label{fig:1}
\end{figure}

The \skweak{} framework (pronounced {\small \textsf{/skwi:k/}}) is a new Python-based  toolkit that provides solutions to this scarcity problem. \skweak{} makes it possible to bootstrap NLP models without requiring any hand-annotated data from the target domain. Instead of labelling data by hand, \skweak{} relies on \textit{weak supervision} to programmatically label data points through a collection of \textit{labelling functions} \cite{fries2017swellshark,Ratner:2017:SRT:3173074.3173077,lison-etal-2020-named,Safranchik_Luo_Bach_2020}. The \skweak{} framework allows NLP practitioners to easily construct, apply and aggregate such labelling functions for classification and sequence labelling tasks.  \skweak{} comes with a robust and scalable aggregation model that extends the HMM model of \citet{lison-etal-2020-named}. As detailed in Section \ref{sec:aggregation}, the model now includes a feature weighting mechanism to capture the correlations that may exist between labelling functions. The general procedure is illustrated in Figure \ref{fig:1}.

Another novel feature of \skweak{} is the ability to create labelling functions that produce \textit{underspecified labels}. For instance, a function may predict that a token is part of a named entity (but without committing to a specific label), or that a sentence does \textit{not} express a particular sentiment (but without committing to a specific class). This ability extends the expressive power of labelling functions and makes it possible to define complex hierarchies between categories -- for instance, {\small \textsf{COMPANY}} may be a sub-category of {\small \textsf{ORG}}, which is itself a sub-category of {\small \textsf{ENT}}. It also enables the expression of ``negative'' signals that indicate that the output should \textit{not} be a particular label. We expect this functionality -- which is to our knowledge absent from other weak supervision frameworks -- to be very useful for NLP applications.

\section{Related Work}
\label{sec:related}

Weak supervision aims to replace hand-annotated `ground truths' with labelling functions that are programmatically applied to data points -- in our case, texts -- from the target domain \cite{Ratner:2017:SRT:3173074.3173077,Ratner2019,lison-etal-2020-named,safranchik:aaai20,fu2020fast}. Those functions may take the form of rule-based heuristics, gazetteers, annotations from crowd-workers, external databases, data-driven models trained from related domains, or linguistic constraints.  A particular form of weak supervision is \textit{distant supervision}, which relies on knowledge bases to automatically label documents with entities  \cite{mintz-etal-2009-distant,ritter-etal-2013-modeling,shang-etal-2018-learning}. Weak supervision is also related to models for aggregating crowd-sourced annotations \cite{pmlr-v22-kim12,hovy-etal-2013-learning,nguyen-etal-2017-aggregating}.  

Crucially, labelling functions do not need to provide a prediction for every data point and may ``abstain'' whenever certain conditions are not met. They may also rely on external data sources that are unavailable at runtime, as is the case for labels obtained by crowd-workers.  After being applied to a dataset, the results of those labelling functions are aggregated into a single, probabilistic annotation layer. This aggregation is often implemented with a generative model connecting the latent (unobserved) labels to the outputs of each labelling function \cite{Ratner:2017:SRT:3173074.3173077,lison-etal-2020-named,Safranchik_Luo_Bach_2020}.  Based on those aggregated labels, a discriminative model (often a neural architecture) is then trained for the task.  

Weak supervision shifts the focus away from collecting manual annotations and concentrates the effort on developing good labelling functions for the target domain. This approach has been shown to be much more efficient than traditional annotation efforts \cite{Ratner:2017:SRT:3173074.3173077}. Weak supervision allows domain experts to directly \textit{inject} their domain knowledge in the form of various heuristics. Another benefit is the possibility to modify/extend the label set during development, which is a common situation in industrial R\&D projects.
 
Several software frameworks for weak supervision have been released in recent years. One such framework is Snorkel \cite{Ratner:2017:SRT:3173074.3173077,Ratner2019} which combines various supervision sources using a generative model. However, Snorkel requires data points to be independent, making it difficult to apply to sequence labelling tasks. Swellshark \cite{fries2017swellshark} is another framework optimised for biomedical NER. Swellshark, is however, limited to classifying already segmented entities, and relies on a separate, ad-hoc mechanism to generate candidate spans. FlyingSquid \cite{fu2020fast} presents a novel approach based on triplet methods, which is shown to be fast enough to be applicable to structured prediction problems. However, compared to \skweak{}, the aggregation model of FlyingSquid focuses on estimating the \textit{accuracies} of each labelling function, and is therefore difficult to apply to problems where labelling sources may exhibit very different precision/recall trade-offs. Finally, \newcite{safranchik:aaai20} describe a weak supervision model based on an extension of HMMs called linked hidden Markov models. Although their aggregation model is related to \skweak{}, they provide a more limited choice of labelling functions, in particular regarding the inclusion of document-level constraints or underspecified labels.

\section{Labelling functions}
\label{sec:lfs}

Labelling functions in \skweak{} can be grouped in four main categories: heuristics, gazetteers, machine learning models, and document-level functions. Each labelling function is defined in \skweak{} as a method that takes SpaCy {\small \textsf{Doc}} objects as inputs and returns text spans associated with labels.\footnote{For classification tasks, the span simply corresponds to the full document itself.}

The use of SpaCy greatly facilitates downstream processing, as it allows labelling functions to operate on texts that are already tokenised and include linguistic features such as lemma, POS tags and dependency relations.\footnote{For languages not yet supported in SpaCy, the multi-language model from SpaCy can be applied.} \skweak{} integrates several functionalities on top of SpaCy to easily create, manipulate, label and store  text documents. 

\subsection*{Heuristics}

The simplest type of labelling functions integrated in \skweak{} are rule-based heuristics. For instance, one heuristic to detect entities of type {\small \textsf{COMPANY}} is to look for text spans ending with a legal company type (such as ``Inc.''). Similarly, a heuristic to detect named entities of the (underspecified) type {\small \textsf{ENT}} is to search for sequences of tokens with NNP POS tags. Section \ref{sec:results} provides further examples of heuristics for  NER and Sentiment Analysis. 

The easiest way to define heuristics in \skweak{} is through standard Python functions that take a SpaCy {\small \textsf{Doc}} object as input and returns labelled spans. For instance, the following function detects entities of type {\small \textsf{MONEY}} by searching for numbers preceded by a currency symbol like \$ or \euro{}:

\begin{lstlisting}[language=Python]
def money_detector(doc):
  """Searches for occurrences of
  MONEY entities in text""" 
  
  for tok in doc[1:]:
      if (tok.text[0].isdigit() and 
          tok.nbor(-1).is_currency):
      yield tok.i-1, tok.i+1, "MONEY"
\end{lstlisting} \vspace{4mm}

\skweak{} also provides functionalities to define heuristics based on linguistic constraints or neighbouring words (within a given window). 
Labelling functions may focus on specific labels and/or contexts. For instance, the heuristic mentioned above to detect companies from legal suffixes will only be triggered in very specific contexts, and abstain from giving a prediction for other text spans. More generally, labelling functions do not need to be perfect and should be expected to yield incorrect predictions from time to time. The purpose of weak supervision is precisely to combine together a set of weaker/noisier supervision signals, leading to a form of denoising \cite{Ratner2019}.

Labelling functions in \skweak{} can be constructed from the outputs of other functions. For instance, the heuristic tagging NNP chunks with the label {\small \textsf{ENT}} may be refined through a second heuristic that additionally requires the tokens to be in title case -- which leads to a lower recall but a higher precision compared to the initial heuristic. \skweak{} automatically takes care of such dependencies between labelling functions in the backend. 

\subsection*{Machine learning models}

Labelling functions may also take the form of machine learning models. Typically, those models will be trained on data from other, related domains, thereby leading to some form of transfer learning across domains. \skweak{} does not impose any constraint on type of model that can be employed. 

The support for underspecified labels in \skweak{} greatly facilitates the use of models across datasets, as it makes it possible to define hierarchical relations between distinct label sets -- for instance, the coarse-grained {\small \textsf{LOC}} label from CoNLL 2003 \cite{tjong-kim-sang-de-meulder-2003-introduction} may be seen as including both the {\small \textsf{GPE}} and {\small \textsf{LOC}} labels in Ontonotes \cite{ontonotes2011}.

\subsection*{Gazetteers}

Another group of labelling functions are \textit{gazetteers}, which are modules searching for occurrences of a list of words or phrases in the document. For instance, a gazetteer may be constructed using the geographical locations from Geonames \cite{wick2015geonames} or names of persons, organisations and locations from DBPedia \cite{journals/semweb/LehmannIJJKMHMK15}

As gazetteers may include large numbers of entries, \skweak{} relies on \textit{tries} to efficiently search for all possible occurrences within a document. A trie, also called a prefix tree, stores all entries as a tree which is traversed depth-first. This implementation can scale up to very large gazetteers with more than one million entries. The search can be done in two distinct modes: a \textit{case-sensitive} mode that requires an exact match and a \textit{case-insensitive} mode that relaxes this constraint.

\subsection*{Document-level functions}

Unlike previous weak supervision frameworks, \skweak{} also provides functionalities to create document-level labelling functions that rely on the global document context to derive new supervision signals. In particular, \skweak{} includes a labelling function that takes advantage of \textit{label consistency} within a document. Entities occurring multiple times through a document are highly likely to belong to the same category \cite{krishnan-manning-2006-effective}. Furthermore, when introduced for the first time in a text, entities are often referred univocally, while subsequent mentions (once the entity is salient) frequently rely on shorter references. \skweak{} provides functionalities to easily capture such document-level relations. 

\section{Aggregation model}
\label{sec:aggregation}
\renewcommand{\UrlFont}{\sffamily\smaller}

After being applied to a collection of texts, the outputs of labelling functions are aggregated using a generative model. For sequence labelling, this model is expressed as a Hidden Markov Model where the states correspond to the ``true'' (unobserved) labels, and the observations are the predictions of each labelling function \cite{lison-etal-2020-named}. For classification, this model reduces to Naive Bayes since there are no transitions. This generative model is estimated using the Baum-Welch algorithm \cite{Rabiner:1990:THM:108235.108253}, which a variant of EM that uses the forward-backward algorithm to compute the statistics for the expectation step. For efficient inference, \skweak{} combines Python with C-compiled routines from the {\small \textsf{hmmlearn}} package\footnote{\url{https://hmmlearn.readthedocs.io/}} for parameter estimation and decoding. 

\subsection{Probabilistic Model}

We assume a list of $J$ labelling functions $\{\lambda_1,..., \lambda_J\}$. Each labelling function produces a label for each data point (including a special ``void'' label denoting that the labelling function abstains from a concrete prediction, as well as underspecified labels). Let $\{l_1,...,l_L\}$ be the set of labels that can be produced by  labelling functions. 

The aggregation model is represented as a hidden Markov model (HMM), in which the states correspond to the true underlying mutually exclusive class labels $\{l_1,...,l_S\}$.\footnote{Note that the set of observed labels $\{\lambda_1,..., \lambda_L\}$ produced by the labelling functions may be larger than the set of latent labels $\{\lambda_1,..., \lambda_S\}$, since those observed labels may also include underspecified labels such as {\scriptsize {\textsf ENT}}.}  This model has multiple emissions (one per labelling function). For the time being, we assume those emissions to be mutually independent conditional on the latent state (see next section for a more refined model). 

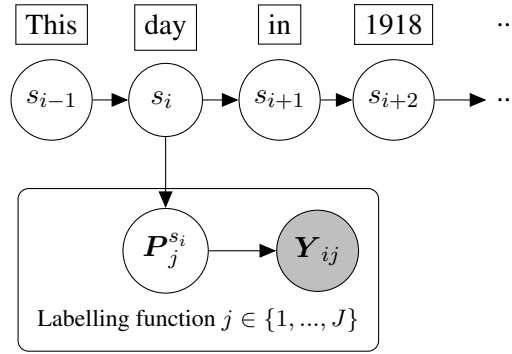
\begin{figure}[t]
\centering
\tikzstyle{token0}=[shape=rectangle]
\tikzstyle{token}=[shape=rectangle, draw=black]
\tikzstyle{alpha}=[shape=circle, draw=black]
\tikzstyle{state}=[shape=circle, draw=black]
\tikzstyle{observation}=[shape=circle,draw=black, fill=lightgray]
\usetikzlibrary{bayesnet}

\begin{tikzpicture}[]
\node[token] (tim) at (0,3) {This} ;
\node[token] (ti) at (1.5,2.96) {day} ;
\node[token] (tip) at (3,3) {in} ;
\node[token] (tipp) at (4.5,3) {1918} ;
\node[token0] (tippp) at (6,3) {...} ;
\node[state] (sim) at (0,2) {$s_{i-1}$} ;
\node[state] (si) at (1.5,2) {$\ s_{i\ } \ $}
 edge [<-] (sim) ;
\node[state] (sip) at (3,2) {$s_{i+1}$}
 edge [<-] (si) ;
\node[state] (sipp) at (4.5,2) {$s_{i+2}$}
 edge [<-] (sip) ;
\node[token0] (sippp) at (6,2) {...}
 edge [<-] (sipp) ;
 \node[alpha] (alphai) at (1.5,0) {$\boldsymbol P_{j}^{s_i}$} 
 edge [<-] (si) ;
 \node[observation] (pij) at (3.5,0) {$\boldsymbol Y_{ij}$} 
 edge [<-] (alphai) ;
 \plate[inner sep=.25cm] {palphai} {(alphai)(pij)} {Labelling function $j \in \{1,...,J\}$} ;
\end{tikzpicture}
\caption{Aggregation model using a hidden Markov model with multiple multinomial emissions.}
    \label{fig:hmm} \vspace{-1mm}
\end{figure}

Formally, for each token $i \in \{1,...,n\}$ and labelling function $\lambda_j$, we assume a multinomial distribution for the observed labels $\boldsymbol{Y_{ij}}$. The parameters of this multinomial are vectors $\boldsymbol{P_j^{s_i}}\in\mathcal{R}^L_{[0,1]}$. The latent states are assumed to have a Markovian dependence structure along the tokens $\{1,...,n\}$. As depicted in Figure~\ref{fig:hmm}, this results in an HMM expressed as a dependent mixture of multinomials:
\begin{align}
 &p(\lambda^{(i)}_{j} = \boldsymbol{Y_{ij}}|\boldsymbol{P^{s_i}_j}) =  \text{Multinomial}\left(\boldsymbol{P_j^{s_i}}\right),\label{themodelbeg}
\\  &p(s_i=k|s_{i-1}=l) = \tau_{lk}.\label{themodeleqend}
\end{align}
where $\tau_{lk}\in\mathcal{R}_{[0,1]}$ are the parameters of the transition matrix controlling for a given state $s_{i-1}=l$ the probability of transition to state $s_{i}=k$.

To initialise the model parameters, we run a majority voter that predicts the most likely latent labels based on the ``votes'' for each label (also including underspecified labels), each labelling function corresponding to a voter. Those predictions are employed to derive the initial transition and emission probabilities, which are then refined through several EM passes. The likelihood function also includes a constraint that requires latent labels to be observed in at least one labelling function to have a non-zero probability. This constraint reduces the search space to a few labels at each step. 

Performance-wise, \skweak{} can scale up to large collections of documents. The aggregation of all named entities from the MUC-6 dataset (see Section \ref{sec:ner}) based on a total of 52 labelling functions only requires a few minutes of computation time, with an average speed of 1000-1500 tokens per second on a modern computing server.

\subsection{Weighting}

One shortcoming of the above model is that it fails to account for the fact that labelling functions may be correlated with one another, for instance when a labelling function is computed from the output of another labeling function. To capture those dependencies, we extend the model with a weighting scheme -- or equivalently, a \textit{tempering} of the densities associated with each labelling function.

Formally, for each labelling function $\lambda_j$ and observed label $k$ we determine weights $\{w_{jk}\}$ with respect to which the corresponding densities of the labelling functions are annealed. This flattens to different degrees the underlying probabilities for the components of the multinomials. 
The observed process has then a tempered multinomial distribution with a  density of form:
\begin{align}
 &p(\lambda^{(i)}_{j} = \boldsymbol{Y_{ij}}|\boldsymbol{P^{s_i}_j},\boldsymbol{w_{j}}) \propto \prod_{k=1}^L{P^{s_i}_{jk}}^{Y_{ijk}w_{jk}}.
\end{align}
The temperatures $\{w_{jk}\}$ are determined using a scheme inspired by delution priors widely used in Bayesian model averaging \cite{george1999discussion, george2010dilution}. The idea relies on  \textit{redundancy} as the measure of prior information on the importance of features. Formally, we define for each $\lambda_j$ a neighbourhood $N(\lambda_j)$ consisting of labelling functions known to be correlated with $\lambda_j$, as is the case for labelling functions built on top of another function's outputs. The weights are then specified as:
\begin{align}
    w_{jk} = \exp{\left(-\gamma   \sum_{l\in N(\lambda_j)}  R_{jlk}\right)},
\end{align}
where $\gamma$ is a hyper-parameter specifying the strength of the weighting scheme, and $R_{jlk}$ is the recall between labelling functions $\lambda_j$ and $\lambda_l$ for label $k$. Informally, the weight $w_{jk}$ of a labelling function $\lambda_j$ producing the label $k$ will decrease if $\lambda_j$ exhibits a high recall with correlated sources, and is therefore at least partially redundant.

\section{Example}

With \skweak{}, one can apply and aggregrate labelling functions with a few lines of code:

\begin{lstlisting}[language=Python]
import spacy, re
from skweak import heuristics, gazetteers, aggregation, utils

# First heuristic (see Section 3)
lf1 = heuristics.FunctionAnnotator         ("money", money_detector)

# Detection of years
lf2= heuristics.TokenConstraintAnnotator ("years", lambda tok: re.match     ("(19|20)\d{2}$", tok.text), "DATE")

# Gazetteer with a few names
NAMES = [("Barack", "Obama"), ("Donald", "Trump"), ("Joe", "Biden")]
trie = gazetteers.Trie(NAMES)
lf3 = gazetteers.GazetteerAnnotator       ("presidents", trie, "PERSON")

# We create a simple text
nlp = spacy.load("en_core_web_md")
doc = nlp("Donald Trump paid $750 in federal income taxes in 2016")

# apply the labelling functions
doc = lf3(lf2(lf1(doc)))

# and aggregate them
hmm = aggregation.HMM("hmm",               ["PERSON", "DATE", "MONEY"])
hmm.fit_and_aggregate([doc])
\end{lstlisting} \vspace{2mm}

\skweak{}'s repository provides Jupyter Notebooks with additional examples and explanations.

\section{Experimental Results}
\label{sec:results}

We describe below two experiments demonstrating how \skweak{} can be applied to sequence labelling and text classification. We refer the reader to \newcite{lison-etal-2020-named} for more results on NER.\footnote{See also \newcite{fries2017swellshark} for specific results on applying weak supervision to biomedical NER.} It should be stressed that the results below are all obtained without using any gold labels.

\subsection{Named Entity Recognition}
\label{sec:ner}

We seek to recognise named entities from the MUC-6 corpus \cite{10.3115/992628.992709}, which contains 318 Wall Street Journal articles annotated with 7 entity types: {\small \textsf{LOCATION}}, {\small \textsf{ORGANIZATION}}, {\small \textsf{PERSON}}, {\small \textsf{MONEY}}, {\small \textsf{DATE}}, {\small \textsf{TIME}}, {\small \textsf{PERCENT}}.

\subsubsection*{Labelling functions}

We apply the following functions to the corpus:
\begin{itemize}\topsep0em\itemsep0em 
\item Heuristics for detecting dates, times and percents based on handcrafted patterns
    \item Heuristics for detecting named entities based on casing, {\small \textsf{NNP}} part-of-speech tags or compound phrases. Those heuristics produced entities of underspecified type {\small \textbf{ENT}}
    \item One probabilistic parser \cite{braun-etal-2017-evaluating} for detecting dates, times, money amounts, percents, and cardinal/ordinal values
    \item Heuristics for detecting person names, based on honorifics (such as Mr. or Dr.) along with a dictionary of common first names
    \item One heuristic for detecting company names with legal suffixes (such as Inc.)
    \item Gazetteers for detecting persons, organisations and locations based on Wikipedia, Geonames \cite{wick2015geonames} and Crunchbase 
    \item Neural models trained on CoNLL 2003 \& the Broad Twitter Corpus \cite{tjong-kim-sang-de-meulder-2003-introduction,derczynski-etal-2016-broad}
    \item Document-level labelling functions based on (1) majority labels for a given entity or (2) the label of each entity's first mention. \vspace{1mm}
\end{itemize}

All together (including multiple variants of the functions above, such as gazetteers in both case-sensitive and case-insensitive mode), this amounts to a total of 52 labelling functions.  

\subsubsection*{Results}

The token and entity-level $F_1$ scores are shown in Table \ref{table:ner}. As baselines, we provide the results obtained by aggregating all labelling functions using a majority voter, along with results using the HMM on various subsets of labelling functions. The final line indicates the results using a neural NER model trained on the HMM-aggregated labels (with all labelling functions). This model is composed of four convolutional layers with residual connections, as implemented in SpaCy \cite{spacy}. See \newcite{lison-etal-2020-named} for experimental details and results for other aggregation methods. 

\begin{table}[t]
\begin{tabular}{lrr} \hline
\multicolumn{2}{l}{Model \hspace{27mm} Token $F_1$} & Entity $F_1$ \\ \hline
Majority vote & 0.64 & 0.62 \\ 
(all labelling functions) & & \\ \hline
HMM-aggregated labels: & & \\
- only heuristics & 0.62 & 0.53 \\
- only gazetteers & 0.46 & 0.39 \\
- only NER models & 0.60 & 0.52 \\
- all but doc-level & 0.83 & 0.74 \\
- all labelling functions & 0.83 & 0.75 \\  \hline
Neural NER trained on & \textbf{0.83} & \textbf{0.76} \\ 
HMM-aggregated labels & & \\ \hline
\end{tabular}
\caption{Micro-averaged $F_1$ scores on MUC-6. } \vspace{-1mm}
\label{table:ner}
\end{table}

\subsection{Sentiment Analysis}

We consider the task of three class (positive, negative, neutral) sentiment analysis in Norwegian as a second case study. We use sentence-level annotations\footnote{Data: \url{ https://github.com/ltgoslo/norec_sentence}} from the NoReC$_{fine}$ dataset \cite{ovrelid-etal-2020-fine}. These are created by aggregating the fine-grained annotations for sentiment expressions such that any sentence with a majority of positive sentiment expressions is assumed to be positive, and likewise with negative expressions. Sentences with no sentiment expressions are labelled neutral. 


\subsubsection*{Labelling functions}

\paragraph{Sentiment lexicons:}
NorSent \cite{barnes-etal-2019-lexicon} is the only available lexicon in Norwegian and contains tokens with their associated polarity. We also use MT-translated English lexicons: 
\textbf{SoCal} \cite{taboada-etal-2011-lexicon}, the \textbf{IBM} Debater lexicon \cite{toledo-ronen-etal-2018-learning} and the NRC word emotion lexicon (\textbf{NRC emo.}) \cite{mohammad-turney-2010-emotions}. Automatic translation introduces some noise but has been shown to preserve most sentiment information \cite{mohammad-etal-translation}.

\paragraph{Heuristics:}

For sentences with two clauses connected by `but', the second clause is typically more relevant to the sentiment, as for instance in ``the food was nice, but I wouldn't go back there''. We include a heuristic to reflect this pattern.

\paragraph{Machine learning models:}

We create a document-level classifier (\textbf{Doc-level}) by training a bag-of-words SVM on the NoReC dataset \cite{velldal-etal-2018-norec}, which contains `dice labels' ranging from 1 (very negative) to 6 (very positive).  We map predictions to positive ($>$4), negative ($<$3), and neutral (3 and 4). We also include two multilingual BERT models \textbf{mBERT-review}\footnote{\url{https://huggingface.co/nlptown/bert-base-multilingual-uncased-sentiment}} (trained on reviews from 6 languages) and \textbf{mBERT-SST} (trained on the Stanford Sentiment Treebank). The predictions for both models are again mapped to 3 classes (positive, negative, neutral). 

\subsubsection*{Results}

\begin{table}[t]
    \centering
    \begin{tabular}{llr}
    \toprule
    & Source & Macro F1 \\
    \cmidrule(lr){2-2}\cmidrule(lr){3-3}
    \multirow{1}{*}{baseline}
    & Majority class     & 22.4 \\   \hdashline
    \multirow{2}{*}{upper bounds}
    & Ngram SVM & 55.2 \\
    & NorBERT & 68.5 \\
    \hdashline
    \multirow{4}{*}{lexicons}
    & NorSent & 45.3 \\
    & NorSent lemmas & 33.7 \\
    & NRC VAD & 8.2 \\
    & SoCal & 46.1 \\
    & SoCal adv. & 43.8 \\
    & SoCal Google & 45.0 \\
    & SoCal Int. & 36.5 \\
    & SoCal verb & 37.2 \\
    & IBM  & 35.9 \\
    & NRC Emo. & 41.7 \\
    \cmidrule(lr){2-2}\cmidrule(lr){3-3}
    \multirow{1}{*}{heuristics}
    & BUT & 25.3 \\
    & BUT lemmas & 24.0 \\
    \cmidrule(lr){2-2}\cmidrule(lr){3-3}
    \multirow{3}{*}{trained models}
    & Doc-level & 33.0 \\
    & mBERT-review & 44.3 \\
    & mBERT-SST & 32.3 \\
    \cmidrule(lr){2-2}\cmidrule(lr){3-3}
    \multirow{2}{*}{Aggregation}
    & Majority vote & 40.0 \\
    & HMM & 49.1 \\
     \cmidrule(lr){2-2}\cmidrule(lr){3-3}
    \multirow{1}{*}{Trained on agg.}
    & NorBERT & \textbf{51.2}\\
    \bottomrule
    \end{tabular}
    \caption{Macro F$_{1}$ on sentence-level NoReC data.} \vspace{-2mm}
    \label{tab:sent_results}
\end{table}

Table \ref{tab:sent_results} provides results on the NoReC sentence test split. As baseline, we include a \textbf{Majority class}  which always predicts the neutral class. As upper bounds, we include a linear SVM trained on TF-IDF weighted (1-3)-grams (\textbf{Ngram SVM}), along with Norwegian BERT (NorBERT) models \cite{kutuzov-etal-2021-norbert} fine-tuned on the gold training data. Those two models are upper bounds as they have access to in-domain labelled data, which is not the case for the other models. Again, we observe that the HMM-aggregated labels outperform all individual labelling functions, and that a neural model (in this case NorBERT) fine-tuned on those aggregated labels yields the best performance.  



\section{Conclusion}

The \skweak{} toolkit provides a practical solution to a problem encountered by virtually every NLP practictioner: how can I obtain labelled data for my NLP task? Using weak supervision, \skweak{} makes it possible to create training data \textit{programmatically} instead of labelling data by hand. The toolkit provides a Python API to apply labelling functions and aggregate their results in a few lines of code. The aggregation relies on a generative model that express the relative accuracy (and redundancies) of each labelling function. The toolkit can be applied to both sequence labelling and text classification and comes along a range of functionalities, such as the integration of underspecified labels. 

\bibliographystyle{acl_natbib}
\bibliography{anthology,acl2021}


\end{document}